\documentclass{article}

\PassOptionsToPackage{numbers}{natbib}
\usepackage[preprint]{neurips_2024}

\usepackage{diagbox}
\usepackage{marvosym}
\usepackage{autobreak}
\usepackage[utf8]{inputenc} 
\usepackage[T1]{fontenc}    
\usepackage{hyperref}       
\usepackage{url}            
\usepackage{booktabs}       
\usepackage{amsfonts}       
\usepackage{nicefrac}       
\usepackage{microtype}      
\usepackage[usenames,svgnames,dvipsnames]{xcolor}
\definecolor{Salmon}{RGB}{228,176,123}
\definecolor{yellowgreen}{RGB}{169,209,142}

\usepackage{graphicx}
\usepackage{multirow}
\usepackage{multicol}

\usepackage{amsmath} 
\usepackage{enumitem}
\usepackage{amssymb}
\usepackage{color}
\usepackage{arydshln}
\usepackage{float}
\usepackage{stfloats}
\usepackage[title]{appendix}
\usepackage{setspace}
\usepackage{subfigure}
\usepackage{verbatim}
\usepackage{comment}
\usepackage{pifont}
\usepackage{natbib}
\title{Perception of Knowledge Boundary for \\ Large Language Models through Semi-open-ended Question Answering}

\author{Zhihua Wen, Zhiliang Tian\thanks{\ \ Corresponding Authors.}, Zexin Jian, {\bf Zhen Huang}, \\{\bf Pei Ke}, {\bf Yifu Gao}, {\bf Minlie Huang}, {\bf Dongsheng Li\footnotemark[1]} \\
College of Computer, National
University of Defense Technology, Hunan, China \\
Tsinghua University, Beijing, China \\
\texttt{\{zhwen, tianzhiliang, gaoyifu, jianzexin21,} \\ \texttt{huangzhen, dsli\}@nudt.edu.cn}, \texttt{aihuang@tsinghua.edu.cn}, \texttt{kepei1106@outlook.com}
}

\begin{document}

\maketitle

\begin{abstract}
Large Language Models (LLMs) are widely used for knowledge-seeking purposes yet suffer from hallucinations. The knowledge boundary of an LLM limits its factual understanding, beyond which it may begin to hallucinate. Investigating the perception of LLMs' \textit{knowledge boundary} is crucial for detecting hallucinations and LLMs' reliable generation. Current studies perceive LLMs' knowledge boundary on questions with a concrete answer (close-ended questions) while paying limited attention to \textit{semi-open-ended questions} that correspond to many potential answers. Some researchers achieve it by judging whether the question is answerable or not. However, this paradigm is not so suitable for semi-open-ended questions, which are usually ``partially answerable questions'' containing both answerable answers and ambiguous (unanswerable) answers. Ambiguous answers are essential for knowledge-seeking, but they may go beyond the knowledge boundary of LLMs. In this paper, we perceive the LLMs' knowledge boundary with semi-open-ended questions by discovering more ambiguous answers. First, we apply an LLM-based approach to construct semi-open-ended questions and obtain answers from a target LLM. Unfortunately, the output probabilities of mainstream black-box LLMs are inaccessible to sample for low-probability ambiguous answers. Therefore, we apply an open-sourced auxiliary model to explore ambiguous answers for the target LLM. We calculate the nearest semantic representation for existing answers to estimate their probabilities, with which we reduce the generation probability of high-probability existing answers to achieve a more effective generation. Finally, we compare the results from the RAG-based evaluation and LLM self-evaluation to categorize four types of ambiguous answers that are beyond the knowledge boundary of the target LLM. Following our method, we construct a dataset to perceive the knowledge boundary for GPT-4. We find that GPT-4 performs poorly on semi-open-ended questions and is often unaware of its knowledge boundary. Besides, our auxiliary model, LLaMA-2-13B, is effective in discovering many ambiguous answers, including correct answers neglected by GPT-4 and delusive wrong answers GPT-4 struggles to identify. 

\end{abstract}

\section{Introduction}
\label{sec:introduction}
Large language models (LLMs) have revolutionized our interactions with AI, enabling users to acquire knowledge by posing questions in natural language~\cite{chowdhery2022palm,NEURIPS2020_1457c0d6}. However, LLMs are prone to hallucination and generate non-factual responses, hindering the development of trustworthy AI. 

    One main cause of LLM hallucination is its unfamiliarity with the long-tail knowledge that appears less frequently than common-sense knowledge in the training data. To alleviate this issue, many researchers collect more domain-specific training data~\cite{mecklenburg2024injecting} or incorporate external information~\cite{wang2023self,ni2024llms} via retrieval-augmented generation (RAG) during inference. Another line of work investigates the perception of \texttt{knowledge boundaries} for LLMs, which indicates the extent of knowledge that the LLM can grasp well, beyond which it may begin to hallucinate~\cite{huang2023survey}.
Studying the perception of knowledge boundaries for LLMs helps alleviate hallucinations in many ways. For example, 1) It helps detect the hallucinations of a target LLM and the extent and scope of its factual knowledge~\cite{hu2024towards,yin2024benchmarking}. 2) It helps align LLMs for more honest generation~\cite{yang2023alignment,xu2024rejection}.

Existing studies on the perception of knowledge boundaries are primarily in the form of Question-Answering (QA). Their methods mainly aim to judge whether a question is answerable or unanswerable, and regard their border as the knowledge boundary. An answerable question refers to when the LLM is capable of generating a response matching the ground truth, and conversely, an unanswerable question means unable to answer correctly. These studies can be categorized into two groups. Prompt-based perception employs prompt engineering~\cite{yin2024benchmarking,deng2024gotcha} to assess whether the LLM can answer the question via LLM self-evaluation. They question whether the LLM knows the answer~\cite{ren2023investigating,cao2023learn,yin2023large} or needs external knowledge to answer the question~\cite{wang2023self}. As LLMs tend to be overconfident~\cite{ni2024llms,ren2023investigating,Huang2023ASO,zhao2023knowing}, more researchers explore representation-based perception. These studies optimize different representations for answers with different answerability~\cite{deng2024gotcha,chen2023adaptation,wang2023self,zhao2023knowing} or extract representations from a fixed encoder to train a classifier~\cite{Khandelwal2020Generalization}.

However, 
directly discriminating questions into answerable and unanswerable ones may not apply to some partially answerable questions. In many scenarios, the questions are relatively open-ended (i.e. having a list of correct answers) that may include (1) a subset of easy answerable answers, and (2) a subset of hard and unpopular answers, which may be unanswerable. These questions (referred to as "semi-open-ended questions") are particularly challenging and knowledge-extensive. Investigating the ambiguous answers to these semi-open-ended questions in various fields benefits knowledge-seeking. Ambiguous answers often go beyond the knowledge boundaries of LLMs and could lead to misinformation (see Sec.~\ref{sec:case_study}). Therefore, we argue that investigating these questions with their ambiguous answers can augment the perception of the knowledge boundaries for LLMs. 

In this paper, we propose to perceive the knowledge boundary for a target LLM with semi-open-ended questions by discovering pieces of unfamiliar knowledge where the LLM learns badly. Particularly, We first construct a dataset with semi-open-ended questions on the open domain and query the target LLM for their corresponding answers. We define the low-probability correct answers and delusive incorrect answers are the ambiguous answers corresponding to the LLM's unfamiliar knowledge. 

A challenge is that obtaining LLMs' low-probability answers needs accessing LLMs' output probabilities (or violently sampling LLMs' outputs many times to approximate the probabilities), which is inaccessible (or expensive) for mainstream black-box LLMs, i.e. GPT-4. Therefore, we approximate the generation probabilities of the target LLM with an open-sourced auxiliary model. We use the Pseudo-inverse of model embedding to estimate the nearest semantic representation for the existing answers. Consequently, we obtain the probability distribution of existing answers and repetitively filter the existing answers (and their semantic-related counterparts) to obtain answers with low-probabilities. Finally, we recognize answers beyond the knowledge boundary of the target LLM by comparing its self-evaluation results against the ground truth answers obtained from RAG-based evaluation. 

Empirically, we use our method to construct a dataset of approximately 1k samples and evaluate GPT-4's performance. We find that GPT-4 makes mistakes in 82.90\% of questions and 40.15\% of its ambiguous answers generated are unqualified. Besides, GPT-4 also makes inaccurate self-evaluation 28.77\% of the time, indicating that these are beyond the knowledge boundary of GPT-4. Moreover, we find nearly 50\% of the candidate answers discovered by our auxiliary model, LLaMA-2-13B, are also beyond the knowledge boundary of GPT-4, including both factual answers that GPT-4 fails to produce and delusive wrong answers GPT-4 evaluates incorrectly.

Our contributions are as threefold:
(1) We are the first to investigate the importance of semi-open-ended questions to the perception of knowledge boundaries for LLMs. 
(2) We propose an ambiguous answer discovery strategy that discovers many ambiguous answers with pieces of knowledge that are beyond the LLM's knowledge boundary.
(3) Experimental results show the poor performance of an advanced LLM, GPT-4, on semi-open-ended questions and the effectiveness of our ambiguous answer discovery method in finding more pieces of knowledge which the LLMs are unfamiliar with. 

\section{Related Work}
\label{sec:related_work}

\subsection{Perception of Knowledge Boundaries for LLMs}
Existing studies on the perception of knowledge boundaries for LLMs can be categorized into prompt-based perception and pattern-based perception. Prompt-based perception perceives the knowledge boundary by querying the target LLM. Many researchers instruct the LLM before and after response generation, asking whether it can correctly answer the questions~\cite{ren2023investigating,wang2023self,cao2023learn,yin2023large} and if the generated answers are accurate~\cite{ren2023investigating,yang2023alignment}. In addition,  \citeauthor{yin2024benchmarking} (\citeyear{yin2024benchmarking}) seek the optimal prompt for benchmarking LLM knowledge boundaries. \citeauthor{amayuelas2023knowledge} (\citeyear{amayuelas2023knowledge}) study the LLMs' ability to understand their knowledge and measure their uncertainty. \citeauthor{kadavath2022language} (\citeyear{kadavath2022language}) also instruct LLMs to generate their confidence score for their responses. As studies find that LLMs tend to be overconfident~\cite{ni2024llms,ren2023investigating,Huang2023ASO,zhao2023knowing}, many researchers explore representation-based perception. Researchers identify unknown questions~\cite{deng2024gotcha} or evaluate correct and incorrect answers~\cite{chen2023adaptation} by implicitly learning their different representations. \citeauthor{chen2023adaptation} (\citeyear{chen2023adaptation}) train LLMs to identify incorrect answers via parameter-efficient tuning. Besides, \citeauthor{wang2023self} (\citeyear{wang2023self}) extract representations of answerable and unanswerable questions to train a classifier to predict whether a question is answerable and assume questions with similar representations share the same answerability. \citeauthor{si2023prompting} (\citeyear{si2023prompting}) take token probability as the answer's confidence score during generation. \citeauthor{zhao2023knowing} (\citeyear{zhao2023knowing}) detect unanswerable questions by paraphrasing questions and checking the divergence of their answer distribution.
The above studies primarily perceive knowledge boundaries for LLMs by distinguishing between answerable and unanswerable questions. This type of binary division does not apply to questions with both common easy answers and unpopular hard answers. Our study is the first to investigate the perception of knowledge boundaries on semi-open-ended questions.

\subsection{Questions Answering for LLMs}

Existing studies on Question Answering (QA) can be categorized into open-ended QA and close-ended QA based on the type of questions. Close-ended questions correspond to a limited number of correct answers, usually in the form of yes or no, true or false, or multiple-choice options, constraining the answers to a predetermined answer set~\cite{rajani-etal-2019-explain,geva-etal-2019-modeling,xiao2021next,robinson2023leveraging}. In addition, Researchers also study open-ended questions that allow the respondent to provide a more detailed and subjective response such as personal opinions and explanations~\cite{karpukhin-etal-2020-dense,xiao2021next,luo2021VQA,bang2023answering}.

Researchers study the performance of LLMs on QA tasks mainly through various prompting strategies. \citeauthor{wei2023chainofthought} (\citeyear{wei2023chainofthought}) explore "Chain-of-Thought" prompting (CoT), a simple and broadly applicable method for enhancing question answering ability of LLMs. \citeauthor{yao2023tree} (\citeyear{yao2023tree}) and \citeauthor{besta2024got} (\citeyear{besta2024got}) introduce similar frameworks for more complex QA tasks, namely, "Tree of Thought" and "Graph of Thoughts" prompting. As studies show that relying solely on an LLM's internal knowledge may lead to hallucinations~\cite{nakano2022webgpt}, many researchers have also improved model performance in QA by incorporating external information (RAG systems~\cite{yue-etal-2023-automatic} and knowledge graph~\cite{2021Zero-Shot, agrawal2024knowledge}). More recently, researchers have studied adaptive retrieval to avoid misinformation in the retrieved documents~\cite{xie2024adaptive}. \citeauthor{ni2024llms} (\citeyear{ni2024llms}) estimate the answerability of the given question and determines whether to retrieve~\cite{chen2023adaptation,ni2024llms}. \citeauthor{xu2024rejection} (\citeyear{xu2024rejection}) learn to identify the knowledge boundaries of LLMs and refuse to answer certain questions to avoid risks~\cite{deng2024gotcha,xu2024rejection}.

\section{Perception of Knowledge Boundary for LLM via Semi-open-ended QA}
\label{gen_inst}
\subsection{Overview}
\begin{figure}[t]
	\centering
	\includegraphics[width=0.98\textwidth]{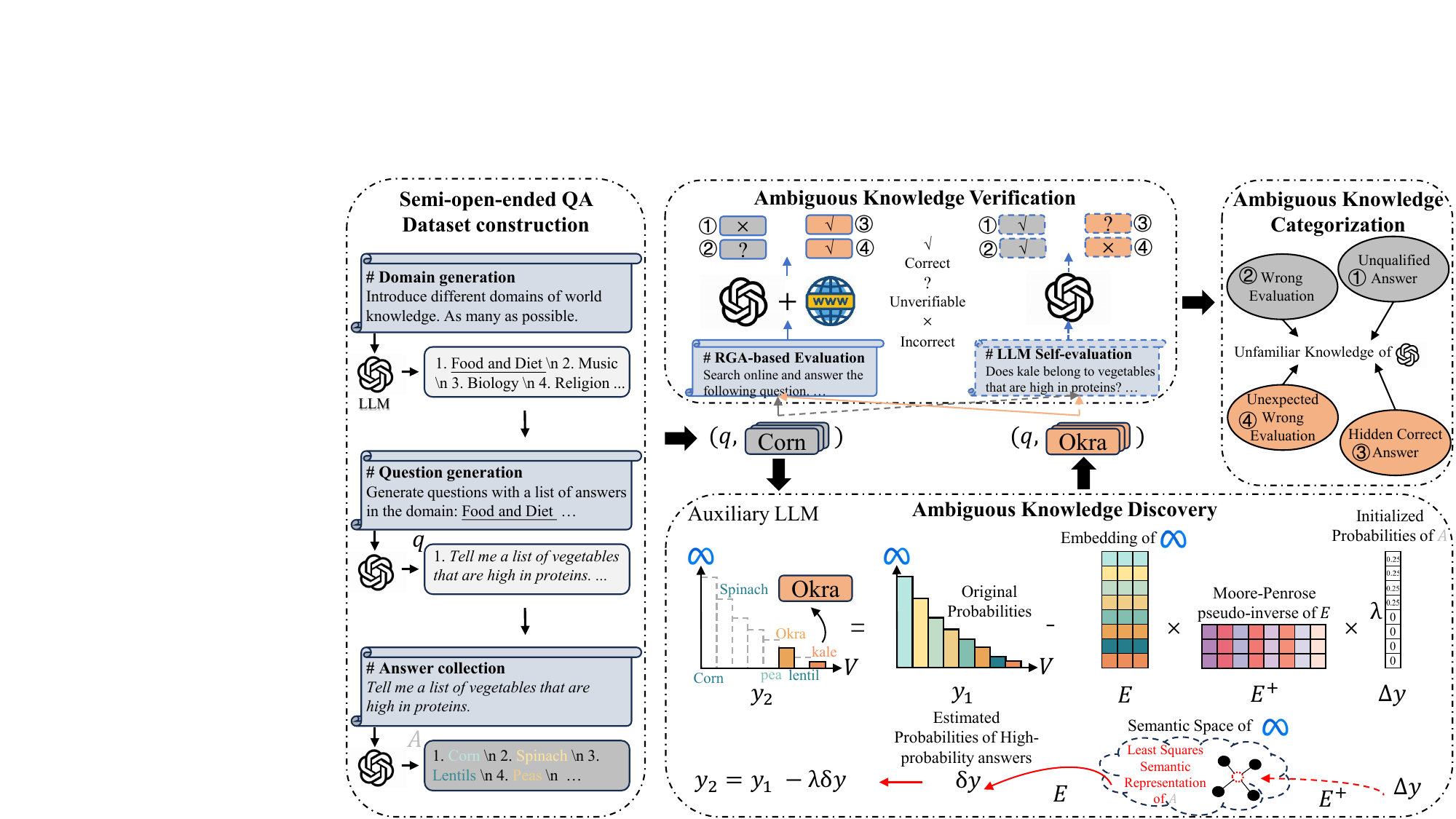}
	\caption{The overview of our framework. For the question $q$ in the constructed dataset, The open-sourced auxiliary model prevent the high-probability answers based on existing answer entities $A$ from the black-box LLM (i.e. GPT-4) and generate 4 categories of ambiguous knowledge that are unfamiliar knowledge for the target model.} 
	\label{fig:main}
\end{figure}

Our framework consists of three parts (see Fig.~\ref{fig:main}). We first exploit the instruction-following ability of a strong LLM to create a dataset consisting of semi-open-ended questions on various domains with LLM's answers. To discover more pieces of unfamiliar knowledge for the target LLM, we apply an open-sourced auxiliary model to incur more ambiguous answers by encouraging more distinctive generations. Finally, we evaluate whether the ambiguous answers to each question are beyond the knowledge boundary of the target LLM by comparing the self-evaluation results against RAG-based evaluation.

\subsection{Semi-open-ended QA Dataset}
\label{}

\subsubsection{Dataset Construction}
\label{sec:dataset_construction}
To study the performance of semi-open-ended questions across various domains, we employ an LLM-based 2-step approach to obtain semi-open-ended questions and collect answers.
\begin{itemize}
\item \textbf{Domain selection.} We first prompt the LLM to generate a list of domains, encompassing world knowledge, which includes areas such as biology, geology, music, etc.
\item \textbf{Question generation.} We prompt the LLM multiple times under each domain to generate a set of semi-open-ended questions $Q$. To ensure the quality of the generation quality, we provide human-written sample questions as demonstrations and specify the following requirements for the generation of candidate questions: 1. The question should correspond to multiple correct answers, making it challenging to answer. The question should also be relatively easy for non-expert users to understand. 2. The judgment of question answers should be based on objective standards in the real world instead of the subjective standards of the evaluator. 3. The truthfulness of an answer to a question should not change constantly over time. 4. The questions share the same template: \textit{Tell me a list of ......}. We use the same vanilla prompt to eliminate the influence of different question styles. 
\item \textbf{Answer collection.} For each question $q$ in $Q$, we query the LLM $I$ times and collect all responses $\textsf{A} =\{a_0,a_1,...,a_{I-1}\}$. In the $i$-th interaction, we inform the LLM of all previously generated responses $\textsf{A}[:i]$ and obtain $a_{i-1}$ by querying the LLM with question $q'$, which repeats the same criteria specified in $q$ and highlights the need for more answers.\footnote{For example, if $q$ is: \textit{Tell me a list of animals unique to Australia.}, then $q'$ is \textit{Tell me more animals unique to Australia.}}. Finally, we extract all answer entities in $\textsf{A}$ and construct an answer list $A$.
\end{itemize}

\subsubsection{Dataset Descriptions}
We create a dataset to investigate the performance of mainstream LLM, GPT-4, on semi-open-ended questions. In dataset construction, we set $I$ to 3 to exploit the knowledge of GPT-4 through multi-round conversations. Like humans, when faced with such questions (e.g., \textit{What are the animals unique to Australia?}), LLMs tend first to give answers in which they hold high confidence (like the \textit{red angaroo}). The latter answers are less certain and may have more mistakes (like \textit{echidna}). We define the initial 75\% of answer entities from GPT-4 generations as common-sense answers, while the remaining 25\% as ambiguous answers. Our dataset comprises 953 questions covering 32 domains, with well-distributed data within each domain. On average, GPT-4 yields 52 answers for each question, including an average of 13 ambiguous answers. See the data samples in App.~\ref{app:dataset}. See the full prompts and demonstrations in App.~\ref{app:instructions}.


\subsection{Ambiguous Knowledge Discovery}
\label{sec:ambiguous_knowledge_sidcovery}
We apply an open-sourced auxiliary LLM to effectively discover more ambiguous answers that may be beyond the knowledge boundaries for black-box LLMs. Our intuition is that low-probability ambiguous answers reflect LLM's unfamiliarity with certain pieces of knowledge. However, it is challenging to collect low-probability ambiguous answers as 1) the generation probability of black-box LLMs (e.g. GPT-4) are inaccessible and their hyper-parameters (e.g. temperature) cannot target ambiguous answer-related tokens (see Sec.~\ref{exp:main}). 2) violently prompting the LLM with question $q$ many times to approximate the generation probability is inefficient as it prioritizes high-probability answers that may already be present in existing answers. 
Answers that are semantically similar to high-probability answers also tend to have a high probability during generation. Hence, we propose to prevent the generation of high-probability answers with their semantic-related counterparts on an open-sourced auxiliary LLM to incur more low-probability ambiguous answers for the perception of the knowledge boundary for the black-box LLM.
We denote that the open-sourced auxiliary model's final layer is $E \in \mathbb{R}^{|V| \times d}$, where $d$ and $V$ is the dimension size and $V$ is the output vocabulary. In each generation step, LLM encodes the semantic representation of the input context as $x \in \mathbb{R}^{d \times 1}$ and uses it to calculate the next token generation probability as $y_1 = Ex$. Specifically, we take the following 3 steps to estimate and decrease the probability of high-probability answers.
\begin{enumerate}
        \item Probability initialization for existing answers. We choose the first token in an answer entity $a \in A$ as ``anchor token'', which is suppose to represent the primary information about $a$ since the first token is indispensable to its generation\cite{chang-etal-2023-multi}. For a given question $q$, we define a vector $\Delta{y} \in \mathbb{R}^{V \times 1}$, indicating the existence of unique anchor tokens in all answer entities to the question. In $\Delta{y}$, we assign the value $\frac{1}{n}$ to the anchor tokens' position and assign 0 to other positions, where $n$ is the number of all anchor tokens. As $\sum_{i} \Delta{y}_i = 1$, we deem $\Delta{y}$ as the initialized probability distribution of all existing answers for the given question.
    \item Semantic estimation of high-probability answers. We estimate the semantic representation $\delta{x}$ for high-probability answer entities from initialized existing answer probability $\Delta{y}$. We  calculate $\delta{x} = E^+ \Delta{y}$, where $E^{+}E = I$. Here, $E^{+}$ is the left Moore-Penrose pseudo-inverse, and $I$ is the identity matrix. 
    Because the pseudo-inverse of a non-square matrix is often used to find the least squares solution of equations, In this way, we obtain $\delta x$ as the least squares semantic representation that is the nearest to approximate the real semantic representation (which is unknown) of anchor tokens.
    \item Probability Reduction. We calculate the probability of high-probability answers as $\delta y = E\delta x$ and ultimately obtain an adjusted generation probability $y_2$:
    $$y_2 = y_1 -\lambda\delta{y} = y_1 - \lambda E\delta{x} = y_1 - \lambda E E_l^+\Delta{y},$$ where $\lambda$ is a scaler that controls the extension of the reduction. Since anchor tokens and their semantic tokens almost share the same semantics\footnote{Near-duplicate tokens roughly share the same semantic meaning but are different under tokenization due to typos, capitalization, or whitespace marking. For example, \textit{Pea} and \textit{peas} are two near-duplicate answers.}, interpreting the estimated semantic representation of anchor tokens propagates the semantic information to their semantic-related ones. Thereby, we obtain an estimated probability distribution of high-probability tokens $\delta y$ and use it to prevent the generation of high-probability answers.
\end{enumerate}
During generation, the auxiliary LLM samples words on the adjusted probability distribution $y_2$ instead of the original distribution $y_1$. In this way, we only reduce the probability of answers that are semantically related to existing high-probability answers. while preserving the remaining probability distribution almost intact. The black-box model has already exhausted almost all high-probability common answers in $A$, forcing the auxiliary model to generate new ambiguous answers with low probabilities. These ambiguous answers are either low-probability correct answers neglected by the black-box model or delusive incorrect answers.



\subsection{Ambiguous Knowledge Verification}
\label{sec:boundary_answer_verify}
We compare the results of LLM self-evaluation against the ground truth from RAG-based evaluation to verify the truthfulness of ambiguous answers, thereby identifying answers beyond the knowledge boundary for the target LLM. We conduct self-evaluation on the target LLM with well-designed instructions. We craft a prompt template including 1) an incentive statement that encourages better performance: \textit{I'll pay you \$100 for a factually correct answer}~\cite{bsharat2024principled}; 2) an instruction \textit{think step by step} which prompts the LLM to analyze before reaching to a conclusion~\cite{NEURIPS2022_9d560961,NEURIPS2022_8bb0d291}. 3) multiple human-crafted examples as in-context learning demonstrations. 

We believe an answer $a$ in each test case $(q,\mathcal{A})$ to be factually correct to $q$ if verified on public, trustworthy sources. As questions in our dataset correspond to numerous low-probability hard answers, it is cost-prohibitive to annotate the truthfulness of each answer with expert knowledge. Inspired by Web-GPT~\cite{nakano2022webgpt}, we adopt a cost-efficient approach to mimic human behavior when faced with unfamiliar knowledge. We instruct an Internet-connected LLM with the same prompt for self-evaluation and require it to search online for related information before making judgments. 

For both evaluations, we evaluate each answer $a$ to its corresponding question as 1) \textit{incorrect}, which contradicts reliable sources; 2) \textit{correct}, which is supported by reliable sources, and 3) \textit{unverifiable}, which is for cases that cannot be verified based on available information. Finally, for each answer $a$, we compare the differences between LLM self-evaluation and RAG-based evaluation to categorize different types of ambiguous knowledge.

\subsection{Ambiguous Knowledge Categorization}
\label{sec:categorization}
We classify the ambiguous answers into four categories based on the above two evaluation results and posit that they are beyond the knowledge boundary of the target LLM. For a question, we categorize the following types of answers to be the LLM's ambiguous answers: 
\begin{itemize}
\item Unqualified answers: answers from the target LLM's response that are identified as incorrect or unverifiable according to the ground truth.
\item Inaccurate evaluations: answers from the target LLM's response whose self-evaluation results contradict their ground truth.
\item Hidden correct answers: answers that are neglected by the target LLM, yet supplemented by the auxiliary model, which are correct according to ground truth.
\item Unexpected wrong evaluations: answers that are neglected by the target LLM but generated by the auxiliary model whose self-evaluation results misalign with their ground truth.
\end{itemize}
The above categorization helps us understand different types of misunderstanding of the target LLM regarding specific pieces of unfamiliar knowledge.

\section{Experiments}
\label{sec::experiments}

\subsection{Settings}
\label{sec::experiments::settings}
In our experiment, we investigate the knowledge boundary of GPT-4 on our constructed dataset. We use two LLaMA-2-13b~\cite{touvron2023llama}, as our auxiliary models in Sec.~\ref{sec:ambiguous_knowledge_sidcovery}. Our method sets $\lambda$ in Sec.~\ref{sec:ambiguous_knowledge_sidcovery} to 80. See more implementation details in App.~\ref{app:eval_detail}. 

We compare our method with several baselines. Following prompt-based approaches~\cite{wang2023self},\textit{Prompt} instructs the auxiliary model to generate more answers via prompt-engineering. Inspired by \citeauthor{zhang2023alleviating} (\citeyear{zhang2023alleviating}), \textit{MASK} belongs to the representation-based perception that uses an average initialization for the probability of tokens from existing answers to represent the likelihood of high-probability answers and reduce their generation probabilities in the auxiliary model.

For the evaluation metrics, we use widely adopted Exact Match~\cite{ren2023investigating} (EM) and F1 scores~\cite{ren2023investigating,wang2023self} to measure the performance of in discovering ambiguous answers. Different from previous research, our semi-open-ended questions correspond to a large number of correct answers, making it hard to build a comprehensive answer set for evaluation\footnote{There are many low-frequency answers for these questions. For example, there are maybe hundreds of accurate answers to the question:\textit{What animals are native to Australia?}}. Instead, we select ambiguous answers identified by the RAG-evaluation and confirmed by our human annotators as their ground truth and compare them with the full response to calculate the \textbf{EM} and \textbf{F1}. In this way, EM is an entity-level metric that measures the percentage of ambiguous answers within the response. F1 is a word-level metric that quantifies the word overlap between the ambiguous answers and the ground truth. We adopt Bleu~\cite{papineni-etal-2002-bleu} to measure word-level overlap between responses from the GPT-4 response and the auxiliary model. We also use answer overlap rate (AOR) to evaluate the efficiency of generating distinctive ambiguous answers. AOR is an entity-level metric that calculates the proportion of words in a list of answer entities that duplicate the reference response.  See more evaluation details in App.~\ref{app:eval_detail}. 


\subsection{Overall Performance}
\label{exp:main}
\begin{table}[htb]
\caption{The performance of different auxiliary models with various strategies in discovering more ambiguous answers.}
\label{tb:main}
\centering
\begin{tabular}{@{}ccccccccc@{}}
\toprule
\textbf{Method}       & \textbf{\begin{tabular}[c]{@{}c@{}}Auxiliary \\ Model Size\end{tabular}} & \textbf{EM $\uparrow$} & \textbf{F1 $\uparrow$} & \textbf{AOR $\downarrow$} & \textbf{Bleu1$\downarrow$} & \textbf{Bleu2$\downarrow$} & \textbf{Bleu3$\downarrow$} & \textbf{Bleu4$\downarrow$} \\ \midrule
Prompt                & LLaMA-2-13B                & 0.300                  & 0.461                  & 0.490                     & 0.252           & 0.118           & 0.052           & 0.023           \\ \midrule
\multirow{2}{*}{MASK} & LLaMA-2-7B                 & 0.458                  & 0.570                  & 0.342                     & 0.185           & 0.075           & 0.021           & 0.004           \\ \cmidrule(l){2-9} 
                      & LLaMA-2-13B                & 0.470                  & 0.587                  & 0.344                     & 0.189           & 0.075           & 0.021           & 0.004           \\ \midrule
Ours                  & LLaMA-2-13B                & \textbf{0.481}         & \textbf{0.587}         & \textbf{0.326}            & \textbf{0.181}           & \textbf{0.071}          & \textbf{0.018}           & \textbf{0.004}           \\ \bottomrule
\end{tabular}
\end{table}
We analyze the effectiveness of our auxiliary model in exploring the knowledge boundary of GPT-4 by comparing it with multiple baselines. We randomly sample 200 questions from our dataset and discover ambiguous answers with an auxiliary model. Then, we verify the truthfulness of these answers using RAG-based evaluation with human annotation following Sec~\ref{sec:boundary_answer_verify} and measure their performance with our evaluation metrics. 
Tab~\ref{tb:main} shows the results of ambiguous answers on different evaluation metrics when using different auxiliary models and strategies. \textit{Prompt} directly prompts the auxiliary model to generate answers, achieving the worst performance on all metrics. This suggests that directly prompting the auxiliary model may generate many repetitive answers (results in high AOR and Bleu scores) and, therefore inefficient in discovering new ambiguous answers (results in low EM and F1). \textit{MASK} reduces the generation probabilities of anchor tokens during generation. When employing \textit{MASK} on the same auxiliary model, its EM and F1 increase to 0.47 and 0.587 respectively while achieving a lower AOR (0.344). It indicates that reducing the anchor words' generation probability is effective in achieving a more diverse generation. Replacing the auxiliary model with LLaMA-7B results in a slightly lower EM of 0.458. This marginal decrease implies that while a larger model can offer a broader knowledge base, the strategy of reducing anchor word probabilities is more influential in generating distinctive answers. Our strategy estimates and reduces the generation probability of near-duplicate answers. This approach achieves the best performance on all metrics. It underscores the effectiveness of our strategy in generating a diverse set of ambiguous answers that are less likely to duplicate existing ones, thus exploring the knowledge boundary for the target LLM. See our case study in Sec.\ref{sec:case_study}.

\subsection{Ablation Study}
\label{exp:rare}

\begin{table}[htb]
\caption{Ablation study on the key components in our method. We use the same metrics as in Sec.~\ref{exp:main}, apart from those that require manual annotation. AOR and Bleu measure the entity- and word-level overlap respectively between answers from GPT-4 and different model variants. $-$ Auxiliary Model prompt GPT-4 with existing answers as examples for more ambiguous answers. $-$ Inverse Matrix keeps the near-duplicate tokens during generation. Besides, we adjust the probability influence scaler ($\lambda$) to verify its impact on the generation results.}
\centering
\label{tb:ablation}
\begin{tabular}{@{}cccccc@{}}
\toprule
\textbf{Variants} & \textbf{AOR}$\downarrow$ & \textbf{Bleu1}$\downarrow$ & \textbf{Bleu2}$\downarrow$ & \textbf{Bleu3}$\downarrow$ & \textbf{Bleu4}$\downarrow$ \\ \midrule
\multicolumn{1}{l}{\textit{$-$ Auxiliary Model}}  & 0.535                                              & 0.267           & 0.106           & 0.037           & 0.010           \\
\multicolumn{1}{l}{\textit{$-$ Inverse Matrix}}  & 0.344                                              & 0.189           & 0.075           & 0.021           & 0.004           \\
\textit{Ours($\lambda$=60)}                & 0.419                                              & 0.224           & 0.098           & 0.038           & 0.013           \\
\textit{Ours($\lambda$=70)}               & 0.352                                              & 0.192           & 0.076           & 0.021           & 0.005           \\ \midrule
\textit{Ours}              & \textbf{0.326}                                     & \textbf{0.181}  & \textbf{0.071}  & \textbf{0.018}  & \textbf{0.004}  \\ \bottomrule
\end{tabular}
\end{table}

We conduct an ablation study on our proposed method to verify the importance of each component in eliciting more distinctive ambiguous answers (as shown in Tab~\ref{tb:ablation}). \textit{$-$ Auxiliary Model} abandons the auxiliary model and use existing answers as in-context learning examples to prompt GPT-4 for more answers. It achieves the highest on all metrics, indicating that prompting the black-box model violently for more answers is inefficient as it results in many repetitive answers. We also try to encourage a more diverse generation by increasing the generation temperature of GPT-4. However, we find that GPT-4 starts generating scrambled texts after just a few words and the perplexity of these texts exceeds 1000, while normally GPT-4 generates a low perplexity of around 10. This indicates that increasing the sampling temperature results in the generation of scrambled texts. Although adjusting the generation temperature can change the generation probabilities, it does not alter the original probability relationships, nor can it specifically target tokens related to ambiguous answers. \textit{$-$ Inverse Matrix} only reduces the probability of existing answers without considering their near-duplicate answers. It performs better than \textit{$-$ Auxiliary Model} while underperforms Ours on all metrics. It shows that estimating and reducing the probability of near-duplicate tokens augment the auxiliary model for higher generation diversity. We increase the intervention on the generation probability by lowering $\lambda$, the probability influence scaler. From row 3 to row 5 in Tab.~\ref{tb:ablation}, $\lambda$ increases from 60 to 80, resulting in a decrease on all metrics. This suggests that the extent to which we intervene in the generation probability is positively correlated with the diversity of ambiguous answers produced by the auxiliary model.


\subsection{Results of Perceiving the Knowledge Boundary for GPT-4}

\begin{table}[htb]
\caption{Percentages of different categories of answers comparing the LLM self-evaluation and ground truth labels. Following the categorization in Sec.~\ref{sec:categorization}, we calculate that the percentage of unqualified answers is 40.15\% by adding up the underlined results that are incorrect or unverifiable according to the ground truth. We also obtain the percentage of inaccurate evaluations as 28.47\%, by adding up the results highlighted in \textcolor{red}{red} where self-evaluation is inconsistent with the ground truth.}
\label{tb:auxiliary_model}
\centering
\begin{tabular}{@{}cccc@{}}
\toprule
\textbf{\begin{tabular}[c]{@{}c@{}}Ground Truth\textbackslash \\ Self-evaluation\end{tabular}} & \textbf{Incorrect}                                                & \textbf{Correct}                                                  & \textbf{Unverifiable}                                             \\ \midrule
\textbf{Incorrect}                                                                                           & \underline{20.98}                                    & \underline{\textcolor{red}{8.37}} & \underline{\textcolor{red}{2.25}} \\ \midrule
\textbf{Correct}                                                                                             & \textcolor{red}{9.30}                          & 47.77                                                             & \textcolor{red}{2.78}                          \\ \midrule
\textbf{Unverifiable}                                                                                        & \underline{\textcolor{red}{3.30}} & \underline{\textcolor{red}{3.73}} & \underline{1.52}                                     \\ \bottomrule
\end{tabular}
\end{table}

By analyzing the ambiguous answers in our dataset, we arrive at the following findings: (1) \textbf{GPT-4 performs poorly on the semi-open-ended questions and generates many unqualified answers.} We calculate the percentage of questions where at least one of the GPT-4's answers is unverifiable or incorrect according to the ground truth. We find that GPT-4 generates incorrect or unverifiable answers in 82.90\% of questions. By adding up the proportion of incorrect answers (row 1 in Tab.\ref{tb:auxiliary_model}) and unverifiable answers (row 3 in Tab.\ref{tb:auxiliary_model}), we identify that 40.15\% of ambiguous answers belong to unqualified answers. (2) \textbf{GPT-4 makes many inaccurate evaluations regarding the truthfulness of ambiguous answers, indicating that the LLM lacks understanding of the relevant knowledge.} We identify answers whose ground truth misaligns with self-evaluations in Tab.\ref{tb:auxiliary_model} and find that 28.47\% ( by adding up the results in red color from Tab.\ref{tb:auxiliary_model}) of ambiguous answers belong to inaccurate evaluations for GPT-4. It indicates that GPT-4 is unfamiliar with these pieces of knowledge, and retrieval is helpful for LLM to draw the correct conclusion. (3) \textbf{GPT-4 has limited ability to recognize its knowledge boundary, while in most cases it continues to produce unqualified answers.} We search for keywords in all responses that reflect GPT-4's admission of its knowledge boundary (e.g. \textit{I apologize} and \textit{I'm afraid}) and calculate the proportion of corresponding questions in the dataset. We find that in about 7\% of questions, GPT-4 admits that it has listed all the answers and refuses to provide more answers (it generates a response like \textit{I apologize for any confusion, but to the best of my knowledge, the list I provided includes all the correct answers.}). However, it fails to recognize its knowledge boundary in the rest questions and continues to generate unqualified answers. Our findings indicate that advanced LLM (i.e. GPT-4) is easy to hallucinate on semi-open-ended questions, indicating the importance of detecting the LLM knowledge boundary via these questions.


\subsection{Results of the Auxiliary Model on Perceiving the Knowledge Boundary for GPT4}
\label{}

\begin{table}[htb]
\caption{Percentages of different categories of ambiguous answers comparing the GPT-4 self-evaluation results and their ground truth. Following the categorization in Sec.\ref{sec:categorization}, we calculate that the percentage of hidden correct answers is 75.12\% by adding up the starred (*) results that are correct according to the ground truth. We also obtain the percentage of unexpected wrong evaluations as 62.43\%, by adding up the results highlighted in \textcolor{orange}{orange} where GPT-4-evaluation is inconsistent with the ground truth.}
\label{tb:auxiliary_model_performance}
\centering
\begin{tabular}{@{}cccc@{}}
\toprule
\textbf{\begin{tabular}[c]{@{}c@{}}Ground Truth\textbackslash \\ GPT-4-evaluation\end{tabular}} & \textbf{Incorrect} & \textbf{Correct} & \textbf{Unverifiable} \\ \midrule
\textbf{Incorrect}                                                                                           & 0.04                  & \textcolor{orange}{9.53}             & \textcolor{orange}{11.31}                  \\ \midrule
\textbf{Correct}                                                                                             &$\textcolor{orange}{23.97}^*$               & $37.54^*$             & $\textcolor{orange}{13.61}^*$
                  \\ \midrule
\textbf{Unverifiable}                                                                                        & \textcolor{orange}{3.18}               &\textcolor{orange}{0.83}             & 0.00                     \\ \bottomrule
\end{tabular}
\end{table}

Tab.\ref{tb:auxiliary_model_performance} shows the fine-grained results of GPT-4 self-evaluation and the ground truth of ambiguous answers discovered by our auxiliary model. By analyzing the results, we conclude some interesting findings: (1) \textbf{LLaMA-2-13B effectively supplements GPT-4 by identifying hidden correct answers.} We add up the starred results in Tab.\ref{tb:auxiliary_model_performance} and obtain the proportion of LLM neglected hidden correct answers (75.12\%). Notably, 23.97\% and 13.61\% of correct ambiguous answers are both neglected by GPT-4 and deemed to be incorrect or unverifiable under GPT-4 self-evaluation, showcasing the GPT-4's unfamiliarity with the corresponding knowledge. (2) \textbf{LLaMA-2-13B is easy to incur unexpected wrong evaluations during GPT-4 self-evaluation.} We add up the results highlighted in orange from Tab.\ref{tb:auxiliary_model_performance} and find that 62.43\% of the GPT-4 self-evaluations are inconsistent with the ground truth. It implies that GPT-4's self-evaluation mechanism may not be fully aligned with the actual correctness of the ambiguous answers, especially when they are supplemented by the auxiliary model. (3) \textbf{LLaMA-2-13B is also able to discover situations where GPT-4 admits for its knowledge boundary.} We add up the results in column 3 where GPT-4 admits that it cannot make a judgment (unanswerable) during self-evaluation and obtain 24.92\% of aligned ambiguous answers. It means that GPT-4 aligns well with these ambiguous answers because it neglected these answers during generation and deems them as unknown knowledge during self-evaluation.

\subsection{Case Study}
\label{sec:case_study}

\begin{table}[htb]
\caption{Examples of two ambiguous answers with their related questions. The Raspberry is an unqualified answer generated by GPT-4, which yields a different answer in another question. Plantain is an answer that is neglected by GPT-4, yet supplemented by the auxiliary model, which are correct according to ground truth. However, GPT-4 believes it to be wrong and generates wrong information for another question. Texts in \textcolor{yellowgreen}{yellowgreen} are truthful information, while texts in \textcolor{red}{red} are non-factual.}
\centering
\large
\setlength\tabcolsep{3pt}
\resizebox{\textwidth}{!}
{
\begin{tabular}{c|cc}
\hline
\textbf{Semi-open-ended Question}                                                               & \multicolumn{2}{c}{Tell me a list of foods that are rich in Vitamin A but low in fat.}                                                                                                                                                                                                                                                                                                                                                                                                                                                                                        \\ \hline
\textbf{\begin{tabular}[c]{@{}c@{}}GPT-4 Response \\ for Semi-open-ended Question\end{tabular}} & \multicolumn{2}{c}{1. Carrots \textbackslash{}n 2. Spinach ... 46. Raspberries \textbackslash{}n 47. Red Leaf Lettuce ...}                                                                                                                                                                                                                                                                                                                                                              \\ \hline
\textbf{\begin{tabular}[c]{@{}c@{}}Auxiliary Model\\ Response\end{tabular}}          & \multicolumn{2}{c}{1. Bell peppers \textbackslash{}n 2. Liver ... 14. Meat such as beef liver \textbackslash{}n 15. Plantains ...}                                                                                                                                                                                                                                                                                                                                                                                                                                                                                                                                 \\ \hline
\textbf{Ambiguous Answer}                                                            & \multicolumn{1}{c|}{Raspberries}                                                                                                                                                                                                                                    & Plantains                                                                                                                                                                                                                                                                    \\ \hline
\textbf{Answer Type}                                                                 & \multicolumn{1}{c|}{Unqualified Answer}                                                                                                                                                                                                                           & Hidden Correct Answer                                                                                                                                                                                                                                                             \\ \hline
\textbf{Related Question}                                                            & \multicolumn{1}{c|}{What vitamins are rich in raspberries?}                                                                                                                                                                                                                 & \begin{tabular}[c]{@{}c@{}}Is it a good choice to eat Plantains for many Vitamin A?\end{tabular}                                                                                                                                                                             \\ \hline
\textbf{GPT-4 Response}                                                              & \multicolumn{1}{c|}{\begin{tabular}[c]{@{}c@{}} \textcolor{yellowgreen}{Raspberries are rich in vitamins C, K, E, and B-complex.} \\ \textcolor{yellowgreen}{They also contain small amounts of vitamin A.} \end{tabular}} & \begin{tabular}[c]{@{}c@{}} \textcolor{red}{Plantains do contain vitamin A,} \\ \textcolor{red}{but not in very high amounts.} ... \end{tabular} \\ \hline
\end{tabular}}
\label{tb:case}
\end{table}

We showcase the importance of ambiguous answers in perceiving the knowledge boundary for LLMs.  Tab.~\ref{tb:case} shows an example with two ambiguous answers for the same question. First, we sample a semi-open-ended question with its GPT-4 responses and ambiguous answers augmented by the auxiliary model (row 2 and 3 in table Tab.\ref{tb:case}). Then, we manually construct two related questions, each involving different types of ambiguous answers, and request new responses from GPT-4. For the case displayed on the left side of Tab.\ref{tb:case}, GPT-4 falsely deems the raspberry as food that is rich in vitamin A but low in fat, yet it answers correctly in another related question. It indicates that GPT-4 is inconsistent in answering different questions involving the same ambiguous answer. For the case on the right side of Tab.\ref{tb:case}, given the same semi-open-ended question, the auxiliary model discovers another ambiguous answer, Plantain, which is correct to the question. Interestingly, GPT-4 generates misinformation regarding this answer entity. It shows that GPT-4 falsely believes plantain is incorrect. It also indicates that the auxiliary model helps discover ambiguous answers that elicit misinformation in the target LLM. It strengthens the importance of perceiving knowledge boundaries for LLMs by discovering ambiguous answers to semi-open-ended questions. See App.~\ref{app:more_cases} for more examples.

\section{Conclusion}
We investigate the perception of knowledge boundary for LLMs with semi-open-ended questions, an important yet underexplored type of question corresponding to a large number of accurate answers. We introduce an LLM-based approach to construct semi-open-ended questions and collect LLM answers from the target LLM. Then, we discover more pieces of unfamiliar knowledge for the target LLM by eliciting ambiguous answers from an auxiliary model that the LLM neglects. To achieve a more effective generation, we estimate and reduce the generation probability of existing answers with their near-duplicate counterparts. With our methods, we construct a dataset to evaluate the performance of GPT-4 and discover many ambiguous answers with our auxiliary model, LLaMA-2-13B. Our findings reveal that GPT-4 produces many unqualified answers and suffers from inaccurate evaluations. Besides, we verify that  LLaMA-2-13B is effective in discovering more unpopular correct answers and delusive wrong answers neglected by GPT-4. Our findings underscore the importance of semi-ended questions and the effectiveness of our method in assisting in perceiving knowledge boundaries for LLMs.

\bibliography{ref}
\bibliographystyle{plainnat}
\clearpage
\appendix
\section{Implimentation Details}
\label{app:eval_detail}
We query GPT-4-Turbo through API calls in our experiments. For our RAG-based evaluation in Sec.~\ref{sec:boundary_answer_verify}, we first employ Microsoft Copilot to search the Internet to find evidence and draw a conclusion. For our evaluation to calculate EM and F1, we hire 11 human annotators with Master's degrees to manually review the responses of Copilot, ensure the credibility of the references, and determine the truthfulness of each answer.

During the generation of our auxiliary model, We apply nucleus sampling (with p=0.9) during generation, setting the generation temperature to 0.7, and the repetition penalty to 1.15.

For our metrics, EM and F1 are common in QA tasks and knowledge boundary detection tasks. The Bleu metric is often used to measure the n-gram overlap between the model response and the ground truth. A lower Bleu score means a more dissimilar response generated by the auxiliary model. Before calculating the above metrics, we normalize answer entities with the NLTK library by lowercase the answers and turning them into the singular form.

\section{Dataset Information}
\label{app:dataset}
\begin{table}[htb]
\caption{Example of data samples in our dataset. Each question corresponds to many ambiguous answers. Different colors of the answers represent different ground truth truthfulness labels: \textcolor{yellowgreen}{yellowgreen} represents that the answer is verified as being factually correct, \textcolor{red}{red} is verified as being incorrect, and \textcolor{Salmon}{salmon} is for unverifiable answers.}
\label{tb:dataset}
\centering
\large
\setlength\tabcolsep{3pt}
\resizebox{\textwidth}{!}
{ 
\begin{tabular}{c|c|c|c}
\hline
\textbf{Domain} & \textbf{Percentage} & \textbf{Question}                                                                                                                                                                                     & \textbf{Ambiguous Answer}                                                                                                                                                                                                                                                                                                                                           \\ \hline
\fontsize{14pt}{17pt}\selectfont Biology           & 4.93\%        & \begin{tabular}[c]{@{}c@{}}Tell me a list of trees that produce fruit with\\ hard shells and are native to tropical regions.\end{tabular}                                                             & \begin{tabular}[c]{@{}c@{}}\textcolor{yellowgreen}{Malabar Chestnut}, \textcolor{red}{Miracle Fruit Tree},\\ \textcolor{yellowgreen}{Guapinol Tree}, \textcolor{red}{Peanut Tree},\\ \textcolor{yellowgreen}{Sapodilla Tree}, \textcolor{yellowgreen}{Desert Date},\\ \textcolor{red}{Castor Oil Plant}, \textcolor{red}{Chiclé Tree},\\ \textcolor{red}{Oriental Persimmon}, \textcolor{yellowgreen}{Dika Tree}, \textcolor{red}{Langsat Tree},\\ \textcolor{yellowgreen}{Pequi Tree}, \textcolor{yellowgreen}{Karite Tree},\\ \textcolor{yellowgreen}{Imbu Tree}, \textcolor{yellowgreen}{Caraipa Tree}\end{tabular} \\ \hline
\fontsize{14pt}{17pt}\selectfont Music             & 4.20\%        & \begin{tabular}[c]{@{}c@{}}Tell me a list of films whose plot is centered on a\\ historically significant event, but the narrative is\\ from the perspective of an imagined protagonist.\end{tabular} & \begin{tabular}[c]{@{}c@{}}\textcolor{yellowgreen}{The Hurt Locker}, \textcolor{red}{Zulu}, \textcolor{red}{Letters from Iwo Jima},\\ \textcolor{yellowgreen}{Doctor Zhivago}, \textcolor{red}{Black Hawk Down}, \textcolor{yellowgreen}{Les Miserables},\\ \textcolor{red}{Lawrence of Arabia}, \textcolor{yellowgreen}{The Wind that Shakes the Barley},\\ \textcolor{red}{The Longest Day}, \textcolor{yellowgreen}{Anthony Adverse},\\ \textcolor{red}{The Guns of Navarone}, \textcolor{red}{Lion of the Desert}\end{tabular}          \\ \hline
\fontsize{14pt}{17pt}\selectfont Geology           & 4.09\%        & \begin{tabular}[c]{@{}c@{}}Tell me a list of current deserts across the world\\ that were previously covered by an ancient sea.\end{tabular}                                                          & \begin{tabular}[c]{@{}c@{}}\textcolor{yellowgreen}{Qaidam Basin}, \textcolor{yellowgreen}{Dasht-e Kavir}, \textcolor{Salmon}{Dasht-e Lut},\\ \textcolor{red}{The Registan Desert}, \textcolor{red}{The Franklin Basin},\\ \textcolor{red}{The Makran Desert}, \textcolor{red}{The Sonoran Desert},\\ \textcolor{red}{The Jornada Del Muerto}, \textcolor{yellowgreen}{Great Karoo}, \textcolor{red}{Tanami Desert}\end{tabular}                      \\ \hline
\fontsize{14pt}{17pt}\selectfont Food and Diet     & 5.67\%        & \begin{tabular}[c]{@{}c@{}}Tell me a list of fruits that are sources of\\ healthy fats, not including avocados and coconuts.\end{tabular}                                                             & \begin{tabular}[c]{@{}c@{}}\textcolor{yellowgreen}{Sacha Inchi Nuts}, \textcolor{yellowgreen}{Hemp Seeds}, \textcolor{red}{Chokeberries},\\ \textcolor{red}{Elderberries}, \textcolor{yellowgreen}{Pine Nuts}, \textcolor{yellowgreen}{Passion Fruit Seeds},\\ \textcolor{yellowgreen}{Tibetan Goji Berries}\end{tabular}                          \\ \hline
\fontsize{14pt}{17pt}\selectfont Literature        & 1.68\%        & \begin{tabular}[c]{@{}c@{}}Tell me a list of female writers from the\\ Victorian era whose work focuses on social reform.\end{tabular}                                                                & \begin{tabular}[c]{@{}c@{}}\textcolor{Salmon}{Helen Taylor}, \textcolor{red}{Catherine Helen Spence}, \textcolor{red}{Octavia Hill},\\ \textcolor{red}{Rhoda Broughton}, \textcolor{red}{Elizabeth Missing Sewell},\\ \textcolor{red}{Emmeline Pankhurst}, \textcolor{yellowgreen}{Alice Meynell}, \textcolor{yellowgreen}{Elizabeth Robins},\\ \textcolor{yellowgreen}{Mary Augusta Ward}, \textcolor{red}{Constance Garnett}\end{tabular}          \\ \hline
\end{tabular}

}
\end{table}

Our dataset covers the following topics: Environment and Climate, Technology and Industry, Political Science, History and Archaeology, Sociology, Economy and Finance, Philosophy, Languages, Art, Architecture, Music, Physics, Astronomy, Chemistry, Biology, Geology, Computer Science, Anthropology and Cultures, Education, Psychology and Mental Health, Fitness and Physical Health, Literature, Religion, Law and Criminology, Military and War
Agriculture, Tourism, Film and Television, Sports and Athletics, Food and Diet, Energy and renewable resources, Mathematics and Statistics, Medicine and Health, Games, Clothing and Fashion.

\section{Important Instructions}
\label{app:instructions}
Here we provide important instructions in building our dataset, guiding LLM for self-evaluation, and show guidelines for human annotators in the supplementary materials.
\begin{itemize}
    \item Domain Generation. \textsl{I hope to test my students' knowledge in different domains. Which domains can I use to create questions?}
    \item Question Generation. \textsl{I am a professor of [CATEGORY] and need to test students' understanding of [CATEGORY] by asking a series of challenging questions. These questions require respondents to list entities that they know meet a series of certain conditions. You need to create more different and diverse challenging questions according to the requirements. Read the following requirements carefully. I'm going to tip \$100 for a perfect list of questions! \textbackslash n  The questions should meet the following criteria: \textbackslash n  1. Each question should start with "Tell me a list of"; \textbackslash n  2. To make the question challenging enough, each question should contain multiple limiting conditions. \textbackslash n  3. The requirement of the question should not involve specific numbers (which makes the question too hard to answer) or vague descriptions (which makes it hard to evaluate the truthfulness of the answer), like "long lifespan", "quick speed", "popular", and "important"; \textbackslash n  4. The boundaries of the question should be very clear, making it easy to evaluate its truthfulness; \textbackslash n  5. The answers to the questions should be consistent through a relatively long time and not change frequently, for example, yearly. \textbackslash n  Refer to the style in the following two examples from an exemplary subject, biology. \textbackslash n  Question 1: Tell me a list of land animals unique to Australia. \textbackslash n  Question 2: Tell me a list of fruits that grow on trees in tropical regions.}
   \item Self-evaluation. \textsl{Does [AMBIGUOUS ANSWER] belong to [QUESTION REQUIREMTNTS]? I'll tip \$100 for the factually correct answer. Think step by step and then give your answer.}
   \item RAG-based evaluation. \textsl{Search online for highly credible information related to the following question, and answer the question based on the search results. \textbackslash n  Does [AMBIGUOUS ANSWER] belong to [QUESTION REQUIREMTNTS]? I'll tip \$100 for the factually correct answer. Think step by step and then give your answer.}
   
\end{itemize}
\section{Cases of Different Types of Ambiguous Answers with Misinformation on the Related Questions}
\label{app:more_cases}

We show different types of ambiguous answers and GPT-4's performance on their related questions from Tab.\ref{app:tab:1} to Tab.\ref{app:tab:2}.

\begin{table*}[htb]
\caption{Examples of two ambiguous answers with their related questions. The Malabo, Equatorial Guinea is an answer that neglected by GPT-4, yet supplemented by the auxiliary model, which are correct according to ground truth. However, GPT-4 believes it to be wrong and generates wrong information for another question. The Rome, Italy is an answer that neglected by GPT-4 but generated by the auxiliary model, whose self-evaluation results misalign with the ground truth. However, GPT-4 believes it to be wrong and generates wrong information for another question. Texts in \textcolor{yellowgreen}{yellowgreen} are truthful information, while texts in \textcolor{red}{red} are non-factual.}
\centering
\label{app:tab:1}
\large
\setlength\tabcolsep{3pt}
\resizebox{\textwidth}{!}
{
\begin{tabular}{c|cc}
\hline
\textbf{Semi-open-ended Question}                                                               & \multicolumn{2}{c}{Tell me a list of world capitals where more than two languages are considered official.}                                                                                                                                                                                                                                                                                                                                                                                                                                                                                        \\ \hline
\textbf{\begin{tabular}[c]{@{}c@{}}GPT-4 Response \\ for Semi-open-ended Question\end{tabular}} & \multicolumn{2}{c}{\begin{tabular}[c]{@{}c@{}} 1. London, United Kingdom \textbackslash{}n 2. Montreal, Canada ... \\ 52. Zagreb, Croatia \textbackslash{}n 53. Prague, Czech Republic ... \end{tabular}}                                                                                                                                                                                                                                                                                                                                                              \\ \hline
\textbf{\begin{tabular}[c]{@{}c@{}}Auxiliary Model\\ Response\end{tabular}}          & \multicolumn{2}{c}{\begin{tabular}[c]{@{}c@{}} 1. Brussels, Belgium \textbackslash{}n 2. Ottawa, Canada ... \\ 9. Rome, Italy \textbackslash{}n 10. Malabo, Equatorial Guinea ... \end{tabular}}                                                                                                                                                                                                                                                                                                                                                                                                                                                                                                                                 \\ \hline
\textbf{Ambiguous Answer}                                                            & \multicolumn{1}{c|}{Malabo, Equatorial Guinea}                                                                                                                                                                                                                                    & Rome, Italy                                                                                                                                                                                                                                                                    \\ \hline
\textbf{Answer Type}                                                                 & \multicolumn{1}{c|}{Hidden Correct Answer}                                                                                                                                                                                                                           & Unexpected Wrong Evaluations                                                                                                                                                                                                                                                             \\ \hline
\textbf{Related Question}                                                            & \multicolumn{1}{c|}{How many official languages are there in Malabo?}                                                                                                                                                                                                                 & \begin{tabular}[c]{@{}c@{}} Is Latin the official language of Rome? \end{tabular}                                                                                                                                                                             \\ \hline
\textbf{GPT-4 Response}                                                              & \multicolumn{1}{c|}{\begin{tabular}[c]{@{}c@{}} There are \textcolor{red}{two} official languages \\ in Malabo: Spanish and Pidgin English. \end{tabular}} & \begin{tabular}[c]{@{}c@{}} \textcolor{red}{No, the official language of Rome is Italian.} ... \end{tabular} \\ \hline
\end{tabular}}
\label{tb:case}
\end{table*}

\begin{table*}[htb]
\caption{Examples of two ambiguous answers with their related questions. The Sand Island Light is an unqualified answer generated by GPT-4, which yields a different answer in another question. The Portsmouth Harbor Light is an answer from GPT-4, whose self-evaluation results contradict the ground truth. However, GPT-4 believes it to be true and generates wrong information for another question. Texts in \textcolor{yellowgreen}{yellowgreen} are truthful information, while texts in \textcolor{red}{red} are non-factual.}
\centering
\label{app:tab:2}
\large
\setlength\tabcolsep{3pt}
\resizebox{\textwidth}{!}
{
\begin{tabular}{c|cc}
\hline
\textbf{Semi-open-ended Question}                                                               & \multicolumn{2}{c}{Tell me a list of famous lighthouses located on islands in the Atlantic Ocean.}                                                                                                                                                                                                                                                                                                                                                                                                                                                                                        \\ \hline
\textbf{\begin{tabular}[c]{@{}c@{}}GPT-4 Response \\ for Semi-open-ended Question\end{tabular}} & \multicolumn{2}{c}{\begin{tabular}[c]{@{}c@{}} 1. Pemaquid Point Light \textbackslash{}n 2. Portland Head Light ... \\ 46. Sand Island Light \textbackslash{}n 47. Portsmouth Harbor Light \textbackslash{}n ... \end{tabular}}                                                                                                                                                                                                                                                                                                                                                              \\ \hline
\textbf{\begin{tabular}[c]{@{}c@{}}Auxiliary Model\\ Response\end{tabular}}          & \multicolumn{2}{c}{\begin{tabular}[c]{@{}c@{}} 1. Maine Coast Guard Station and Pier Light \textbackslash{}n 2. Minot's Leading Light ... \\ 10. Woodenboat School Light \textbackslash{}n 11. Chappaquiddick Light ... \end{tabular}}                                                                                                                                                                                                                                                                                                                                                                                                                                                                                                                                 \\ \hline
\textbf{Ambiguous Answer}                                                            & \multicolumn{1}{c|}{Portsmouth Harbor Light}                                                                                                                                                                                                                                    & Sand Island Light                                                                                                                                                                                                                                                                    \\ \hline
\textbf{Answer Type}                                                                 & \multicolumn{1}{c|}{Inaccurate Evaluations}                                                                                                                                                                                                                           & Unqualified Answer                                                                                                                                                                                                                                                             \\ \hline
\textbf{Related Question}                                                            & \multicolumn{1}{c|}{Is Portsmouth Harbor Light located on which island in the Atlantic Ocean?}                                                                                                                                                                                                                 & \begin{tabular}[c]{@{}c@{}} Where is Sand Island Light located? \end{tabular}                                                                                                                                                                             \\ \hline
\textbf{GPT-4 Response}                                                              & \multicolumn{1}{c|}{\begin{tabular}[c]{@{}c@{}} \textcolor{red}{Portsmouth Harbor Light is located on New Castle Island}\\ \textcolor{red}{in the Atlantic Ocean, in the state of New Hampshire, USA.} \end{tabular}} & \begin{tabular}[c]{@{}c@{}} \textcolor{yellowgreen}{Sand Island Light is located in Alabama, USA,} \\ \textcolor{yellowgreen}{specifically on Sand Island off}  \\ \textcolor{yellowgreen}{the coast of Dauphin Island.} ... \end{tabular} \\ \hline
\end{tabular}}
\label{tb:case}
\end{table*}

\end{document}